\documentclass[sigconf, nonacm]{acmart} 

\usepackage{dsbda-style}
\usepackage{graphicx}
\graphicspath{ {./images/} }
\usepackage{comment}
\usepackage{textgreek}
\usepackage{diagbox}
\usepackage{underscore}
\usepackage{hyperref}

\usepackage{subcaption}
\usepackage{placeins}

\usepackage{multirow}

\AtBeginDocument{\providecommand\BibTeX{{\normalfont B\kern-0.5em{\scshape i\kern-0.25em b}\kern-0.8em\TeX}}}

\begin{document}

\title[Text Role Classification of Scientific Charts Using Multimodal Transformers]{Text Role Classification in Scientific Charts \\ Using Multimodal Transformers}

\author{Hye Jin Kim}
\email{hye.kim@alumni.uni-ulm.de}
\orcid{0009-0007-9582-2670}
\affiliation{\institution{Universität Ulm}
  \city{Ulm}
  \country{Germany}
}

\author{Nicolas Lell}
\email{nicolas.lell@uni-ulm.de}
\orcid{0000-0002-6079-6480}
\affiliation{\institution{Universität Ulm}
  \city{Ulm}
  \country{Germany}
}

\author{Ansgar Scherp}
\email{ansgar.scherp@uni-ulm.de}
\orcid{0000-0002-2653-9245}
\affiliation{\institution{Universität Ulm}
  \city{Ulm}
  \country{Germany}
}

\renewcommand{\shortauthors}{Kim, Lell, Scherp}

\begin{abstract}

Text role classification involves classifying the semantic role of textual elements within scientific charts. 
For this task, we propose to finetune two pretrained multimodal document layout analysis models, LayoutLMv3 and UDOP, on chart datasets. 
The transformers utilize the three modalities of text, image, and layout as input. 
We further investigate whether data augmentation and balancing methods help the performance of the models. 
The models are evaluated on various chart datasets, and results show that LayoutLMv3 outperforms UDOP in all experiments. 
LayoutLMv3 achieves the highest F1-macro score of $82.87$ on the ICPR22 test dataset, beating the best-performing model from the ICPR22 CHART-Infographics challenge. 
Moreover, the robustness of the models is tested on a synthetic noisy dataset ICPR22-N. 
Finally, the generalizability of the models is evaluated on three chart datasets, CHIME-R, DeGruyter, and EconBiz, for which we added labels for the text roles. 
Findings indicate that even in cases where there is limited training data, transformers can be used with the help of data augmentation and balancing methods. The source code and datasets are available on GitHub: \url{https://github.com/hjkimk/text-role-classification}.

\end{abstract}


\maketitle

\section{Introduction}

Writing a scientific paper is a long and tedious process. There are many systems available to edit the textual content of papers such as tools to check spelling and grammar, including Grammarly\footnote{\url{https://www.grammarly.com}}. Similarly, some tools check the image content of scientific papers (e.\,g., scientific figures) including JetFighter\footnote{\url{https://elifesciences.org/labs/c2292989/jetfighter-towards-figure-accuracy-and-accessibility}}, which examines figures to improve data representation, and Barzooka\footnote{\url{https://github.com/quest-bih/barzooka}}, which detects various chart types.

Scientific figures are an important part of scientific papers for visualizing information. Therefore, a feedback system dedicated to scientific figures may save time and also improve the overall comprehensibility of a paper. One function of such a system that could support the authors is checking whether scientific charts have the necessary elements (e.\,g., title, axis labels, axis ticks, and legend). In some cases, authors forget to include these elements which can hamper the readability of charts. In the literature, this task is called text role classification~\cite{DavilaKSGTSC19, DavilaTSSSG20, DavilaXAMSG22, Al-ZaidyG17, WangCZX20, WuXHTLWCLCLDHJ21}. It aims to identify the semantic role of the text objects embedded in charts. 
This task can be incorporated into a feedback system for scientific papers, which authors can use to check their papers before submission.

There is only few research on text role classification in scientific charts. 
Although there exists the Challenge on Harvesting Raw Tables from Infographics (CHART-Infographics), in which one of the tasks is dedicated to text role classification~\cite{DavilaKSGTSC19, DavilaTSSSG20, DavilaXAMSG22}, there are no publications of the best-performing methods used in the competition. 
Outside of the competition, there have only been a few publications on text role classification~\cite{WangCZX20, PocoH17, YanAD23}. 
Furthermore, while the performance of text role classification models on synthetically created datasets is near perfect, there is a drastic drop in performance on real datasets consisting of charts extracted from scientific publications.

\begin{figure}[h]
  \centering
  \includegraphics[width=\linewidth]{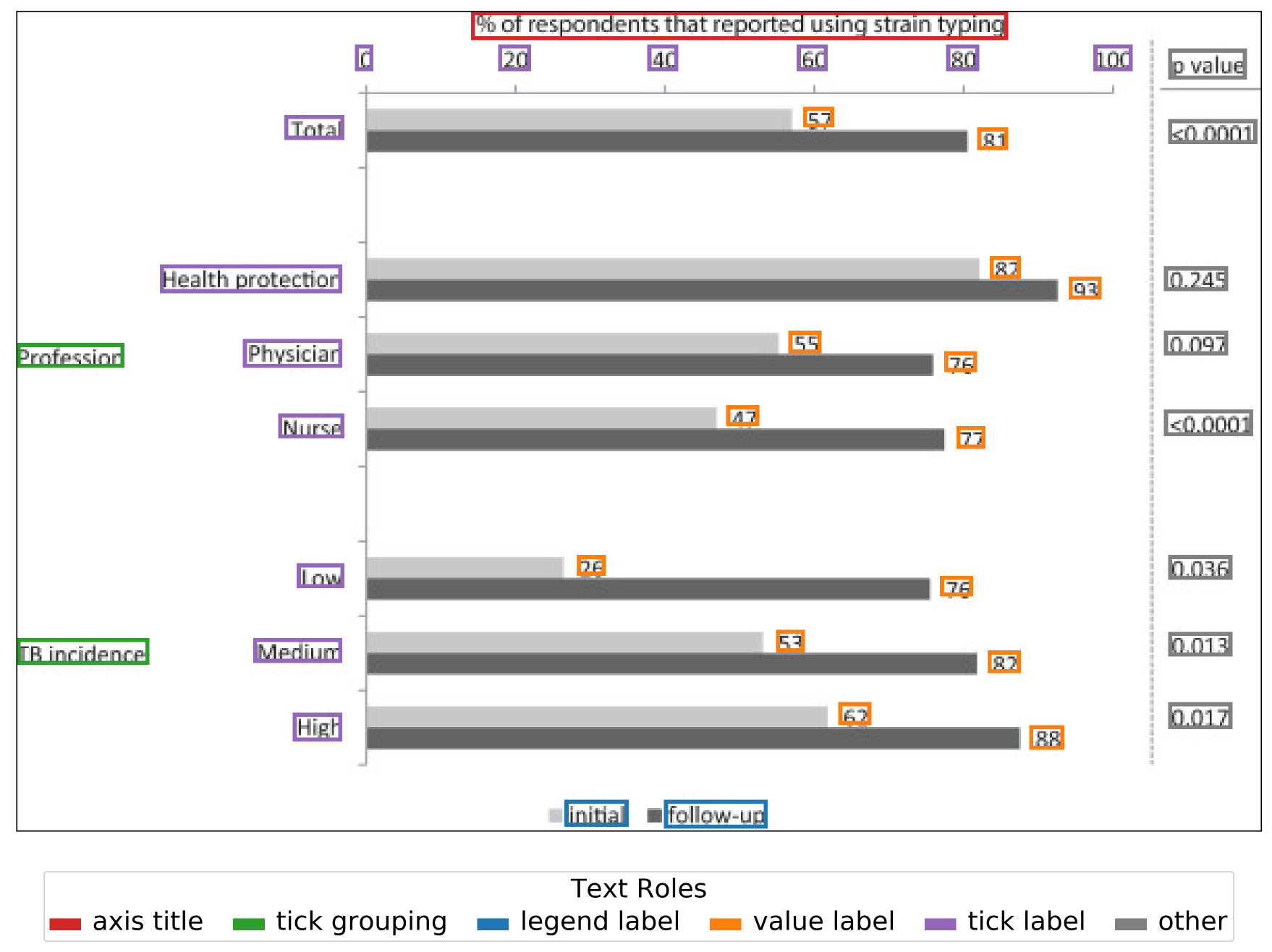}
  \caption{A sample bar chart from ICPR22. Along with the chart image and the text, the text bounding box coordinates are used as the position modality for the multimodal input to the transformers.}
  \label{fig:figure-1}
\end{figure}

The motivation of this work is to improve the performance of text role classification by using transformers with multimodal data, namely the chart image\footnote{For clarification, the term ``image'' in this paper refers to an image of a single scientific chart. 
Among the various terms that exist, charts~\cite{DavilaKSGTSC19, DavilaTSSSG20, DavilaXAMSG22, Al-ZaidyG17} and figures~\cite{BoschenBS18, JessenBS19, ChoudhuryG15} are commonly used in the literature. 
To be consistent with the work on CHART-Infographics, the term ``chart'' will be used in this paper.}, the text elements within the chart, and their position. 
An example of this multimodal chart data can be found in Figure \ref{fig:figure-1}. More specifically, we examine whether using multimodal transformers designed for and pretrained on large document datasets (i.\,e., document analysis models) can be finetuned for charts. In addition, we investigate whether data augmentation and balancing techniques improve the performance of these models. Moreover, there are no prior experiments measuring the generalizability of the models to new chart data and the robustness of text role classification models to noise. Therefore, the models are evaluated on three chart datasets with added labels for the text role classes CHIME-R, DeGruyter, and EconBiz. The robustness of the models is also analyzed on a noisy chart dataset ICPR22-N.
In summary, our contributions are:
\begin{itemize}
  \item We finetune two top-performing document layout analysis models LayoutLMv3 and UDOP for text role classification on either ICPR22 only or on all chart datasets combined. In both cases, LayoutLMv3 outperforms UDOP on ICPR22. Comparing the performances of the models to two baseline models from the ICPR 2022 CHART-Infographics competition~\cite{DavilaXAMSG22} and CACHED~\cite{YanAD23}, LayoutLMv3 finetuned on only ICPR22 achieves the highest F1-macro score of 82.87 and an F1-micro score of 93.99 while UDOP outperforms only one method.
  \item Data augmentation and balancing methods are used to increase the performance of the models. While applying data augmentation and balancing improves the F1-micro and macro score on ICPR22 for UDOP, there is only a minor improvement of the F1-micro score for LayoutLMv3.
  \item We evaluate the generalizability of the models on CHIME-R, DeGruyter, and EconBiz. LayoutLMv3 outperforms UDOP on all three datasets. For both models, finetuning on all chart datasets with data augmentation and balancing generally results in the best F1 scores. The highest F1 scores are on CHIME-R, and the lowest are on DeGruyter. Furthermore, the robustness of the models is assessed on ICPR22-N, revealing that LayoutLMv3 is more robust to noise compared to UDOP. While data augmentation and balancing methods improve the model robustness of UDOP on ICPR22-N, it does not do so for LayoutLMv3.
  \item Three chart datasets CHIME-R, DeGruyter, and EconBiz are labeled to include text role class information for each text element. In addition, a synthetic but realistic noisy dataset ICPR22-N is created.  
\end{itemize}

In the following section, we summarize the related work. In Section 3, the two document layout analysis models LayoutLMv3 and UDOP are described in more detail, and the data augmentation and balancing methods are introduced. The datasets and experimental procedure are specified in Section 4 and the results are reported in Section 5. We discuss the results in Section 6 and reflect on the limitations of the experiments before we conclude.

\section{Related Work}
\label{sec:relatedwork}
We introduce works on unimodal transformers, multimodal models, and text role classification. The two layout analysis models used in this paper for text role classification, LayoutLMv3~\cite{HuangL0LW22} and UDOP~\cite{TangYWFLZZZB}, are multimodal transformers. Therefore, we first explain the basic unimodal transformers, and then introduce various multimodal models including multimodal transformers. Finally, the details of text role classification are described.

\subsection{Unimodel Transformers}
Transformers~\cite{VaswaniSPUJGKP17} are commonly used in NLP because of their high performance. A self-attention mechanism in the encoder-decoder architecture is used to compute representations of a sequence by relating word tokens at different positions in the input sequence. This allows the model to understand text with its context while being parallelizable. BERT~\cite{DevlinCLT19} is an encoder-only architecture that introduced bidirectional training to transformers. Two pretraining tasks Masked Language Model (MLM) and Next Sentence Prediction (NSP) are used to train BERT. The objective of MLM is to predict masked tokens based on the left input, and the objective of NSP is to predict if sentence pairs are related in terms of whether the second sentence is successive to the first sentence. BERT can then be finetuned on downstream tasks such as text classification.

With the introduction of ViT~\cite{DosovitskiyB0WZ21}, transformers showed to be effective for vision tasks as well. Instead of word tokens, patches of images are used as tokens. A sequence of linear embeddings of flattened image patches along with the positional embeddings are fed as input to the transformer encoder. When trained on large datasets, ViT outperforms CNNs on image classification tasks. Even for smaller datasets, ViT shows competitive results in comparison to models trained on large datasets. To achieve this,~\citet{SteinerKZWUB21} suggested training longer as well as applying data augmentation and regularization. In the experiments by~\citet{SteinerKZWUB21}, MixUp~\cite{ZhangCDL18} and RandAugment~\cite{CubukZSL20} are used for data augmentation, and dropout and stochastic depth-regularization are applied for regularization.~\citet{TouvronCDMSJ21} used a teacher-student strategy by adding a distillation token with which the model tries to predict the teacher label.

A lot of the research on transformers is dedicated to enhancing various aspects such as generalization and complexity. To improve the generalization of transformers, CoAtNet~\cite{DaiLLT21} merges convolution and self-attention in one computational block. Swin Transformer~\cite{LiuL00W0LG21} is a hierarchical vision transformer using shifted windows that allows the model to have linear computational complexity with respect to the image size, compared to the quadratic computational complexity of previous transformer-based architectures. In fact,~\citet{TolstikhinHKBZU21} showed that MLP-Mixer, a simple transformer architecture consisting of only Multilayer Perceptrons (MLP), is an equally performing alternative to convolutions and attention mechanisms.

\subsection{Multimodal Models}
While BERT and ViT are unimodal models, the nature of the text role classification task is multimodal. Due to the limitations of using unimodal models for multimodal tasks, there has been a growing interest in the field of multimodal models to bridge the gap between NLP and computer vision. An example of a multimodal task found in the literature is image-text classification where the objective is to classify image-text pairs into their correct label~\cite{KielaBFT19, RadfordKHRGASAM21, FuXLLXWLSW22}. MMBT~\cite{KielaBFT19} achieves this task by concatenating image embeddings from linear projections of ResNET~\cite{HeZRS16} to BERT text embeddings. In addition to position embeddings, a segment embedding is used to inform the multimodal transformer which part of the input vector corresponds to the text and which to the image.

More recent multimodal models avoid reliance on convolutions. CLIP~\cite{RadfordKHRGASAM21} uses separate transformer encoders for the image and text. The image and text encoders are jointly trained to predict the correct pairings, maximizing the cosine similarity of the corresponding image and text embeddings. CMA-CLIP~\cite{FuXLLXWLSW22} uses CLIP as backbone in addition to two attention modules. First, the sequence-wise attention module concatenates image and text embeddings. These are given as input into the transformer, which outputs an aggregated embedding for both image and text. Subsequently, a modality-wise attention module is used to learn the weighted sum of the two aggregated embeddings based on how relevant each modality is to the downstream task.

For text role classification, the layout of the chart is important to consider because there are common assumptions of the layout based on the chart type. For example, line and bar charts usually have axis labels while pie charts do not. 
One domain in visual language understanding that is related to chart analysis is document layout analysis. 
Charts and documents both contain text and images and usually have some sort of structure. 
In fact, charts may be embedded in documents. LayoutLM~\cite{XuL0HW020} is a BERT-like pretrained model that jointly learns text and layout information obtained from the bounding box coordinates of the text. A separate image model is used to extract the image embeddings, which are combined with the LayoutLM embeddings to perform downstream tasks during finetuning. LayoutLMv2~\cite{XuXL0WWLFZCZZ20} is an improvement of LayoutLM, where image embeddings from a CNN-based image encoder are concatenated with the text embeddings during pretraining. For the layout information, 1D position, 2D position, and segment embeddings are provided. LayoutLMv3~\cite{HuangL0LW22} mitigates the use of CNNs for extracting visual features by using linear embeddings of flattened image patches. Image embeddings are concatenated with text embeddings and the transformer is provided with spatial information from the 1D and 2D position embeddings. LayoutLMv3 is pretrained using Masked Language Modeling (MLM) for text, Masked Image Modeling (MIM) which is a similar technique to MLM for images, and Word-Patch Alignment (WPA) to learn the correspondence between the text and image embeddings. Instead of using a pretraining objective to learn the layout information, UDOP~\cite{TangYWFLZZZB} is a sequence-to-sequence generative transformer that uses a layout-induced vision-text embedding by calculating the joint representation of image patches that contain text. This means that UDOP only employs one encoder to represent the multimodal input. However, it uses two different decoders, one for vision and another for text-layout to generate outputs.

\subsection{Text Role Classification}
\label{sec:textroleclassification}

Text role classification is comparable to document region classification~\cite{Bhowmik23}, a document layout analysis task that aims to classify the role of segmented regions of documents. 
The similarity between the two tasks suggests that pretrained document layout analysis models can be finetuned for text role classification. 
Nonetheless, it should be mentioned that there are two main differences between text role classification and document region classification. First, while text role classification classifies the roles of text specifically within charts, document layout analysis classifies the roles of document regions that are not necessarily text (e.\,g., tables and images). Second, although text role classification can be performed on documents that contain charts, document layout analysis cannot be done on charts.

Early work on text role classification applied heuristic-based methods using a combination of classifiers based on Support Vector Machines (SVM), Random Forests, Decision Trees, Naive Bayes, and object detection networks with geometric features, layout features, and text-based features~\cite{DavilaSDKG21}. 
The disadvantages of heuristic-based approaches is that they are not robust to different chart types and layouts and are difficult to maintain. Therefore, recent approaches implement neural network-based models.

It is important to have standardized benchmark datasets to compare existing methods. The Challenge on Harvesting Raw Tables from Infographics (CHART-Infographics) provides such benchmark datasets. The overall goal of the challenge is to create an end-to-end process for extracting data from charts. This process is divided into seven individually evaluated tasks, one of which is text role classification. To date, three benchmark datasets for text role classification are available from three different years of CHART-Infographics: ICDAR 2019~\cite{DavilaKSGTSC19}, ICPR 2020~\cite{DavilaTSSSG20}, and ICPR 2022~\cite{DavilaXAMSG22}. The datasets from the challenges are called ICDAR19, ICPR20, and ICPR22 respectively. For both ICPR20 and ICPR22, each text element is classified into nine roles: chart title, legend title, legend label, axis title, tick label, tick grouping, mark label, value label, and other. For ICDAR19, only four role classes were considered: chart title, axis title, tick label, and legend label. While ICDAR19 and ICPR20 include synthetic charts as well as real charts, ICPR22 only contains real charts. This was done to encourage the improvement of model performance on real chart datasets because in the previous competitions, although the performance on synthetic chart datasets was always near perfect, the performance on real chart datasets was drastically lower.

Heuristic-based approaches are used in the top-performing submissions for the CHART-Infographics from ICDAR 2019 while neural network-based approaches are used in the best submissions in ICPR 2020 and ICPR 2022. 
ABC, the top-performing submission from ICDAR 2019\footnote{\url{https://chartinfo.github.io/index\_2019.html}}, implemented a gradient boosting decision tree, which was trained on 20 heuristic features of the bounding boxes and text. 
This includes the aspect ratio, position, and number of horizontally/vertically aligned bounding boxes, as well as information on whether the text is numeric or not, text orientation, and relative position of the text with respect to the axis line or legend~\cite{DavilaKSGTSC19}. In ICPR 2020\footnote{\url{https://chartinfo.github.io/leaderboards\_2020.html}}, the top-performing submission, Lenovo-SCUT-Intsig, applied a weighted ensemble of three models using text, image, and position features as well as text semantics and chart type~\cite{DavilaTSSSG20}. \citet{WuXHTLWCLCLDHJ21} proposed a method that was not submitted to the challenge but outperformed the best methods from ICDAR 2019 and ICPR 2020. The method fused the extracted visual features with context features. Visual features are extracted using a ResNet backbone with RoIAlign~\cite{HeGDG17}, which aligns the extracted features from each Region of Interest (RoI) with the input. Context features are extracted using LayoutLM~\cite{XuL0HW020} using word-level embeddings and multimodal embeddings (including position and chart type). After fusing the features, a self-attention module is applied, and finally, a fully-connected layer outputs the text role classification results.

IIIT CVIT Chart Understanding, the winning submission from ICPR 2022\footnote{\url{https://chartinfo.github.io/leaderboards\_2022.html}}, proposed a Cascade Mask R-CNN~\cite{HeGDG17} with Swin Transformer as backbone~\cite{DavilaXAMSG22}. 
The second best submission, UB-ChartAnalysis, used a similar Cascade R-CNN~\cite{CaiV18} that was modified for understanding context as well as the local and global information~\cite{DavilaXAMSG22}. 
Apart from the ICPR 2022 submissions, CACHED~\cite{YanAD23} is another chart understanding model that outperforms UB-Chart\-Analysis on text role classification. 
CACHED is a Cascade R-CNN framework with Swin Transformer as backbone, which focuses on fusing the local and global context by enhancing visual context and encoding positional context. 
Visual context is enhanced by incorporating global features into RoI, and positional context is encoded by concatenating bounding box features with RoI vision features. 

\section{Methods}
\label{sec:methods}

The related work suggests that using multimodal transformers with features that include text, image, and position information can produce favorable outcomes for text role classification. 
However, training transformers from scratch is computationally expensive and chart data with text role annotations is limited. 
Thus, it is worth investigating whether pretrained multimodal transformers can be finetuned for text role classification, despite being pretrained on non-chart datasets. 
We chose LayoutLMv3 and UDOP for experimentation because they are the top-performing multimodal transformers for document layout analysis. 

In this section, we first describe in more detail the two pretrained models LayoutLMv3 and UDOP, focusing on the difference between how the multimodal inputs are embedded. We then explain the data augmentation methods applied to increase the robustness of the models, and the data balancing methods used to account for imbalanced datasets.

\subsection{Models}

\paragraph{LayoutLMv3}

This multimodal transformer uses the tokenizer from RoBERTA~\cite{LiuOGDJCLLZS19} to generate text embeddings with Byte-Pair Encoding (BPE). 
For the image embeddings, the image tokenizer from DiT~\cite{LiXLCZW22} is used which consists of a discrete variational autoencoder (dVAE) from DALL-E~\cite{RameshPGGVRCS21}, trained on a large-scale document image dataset. 
The position embeddings consist of 1D text positions and 2D positions containing the bounding box coordinates. 
To learn the layout information of the image, the pretraining objective Word-Patch Alignment (WPA) is used to predict whether image patches containing text are masked. 
The base model of LayoutLMv3 adopts a 12-layer transformer encoder with 12 self-attention heads, a hidden size of $D = 768$, and a feed-forward network with one hidden layer of size 3,072.
A classifier head is added to the encoder output. 
Excluding the classifier parameters, it has in total 133M trainable parameters.

\paragraph{UDOP}

While LayoutLMv3 uses a pretraining objective to learn layout information, UDOP uses layout-induced vision-text embeddings for a joint representation of the image and text. 
The tokenizer from T5~\cite{RaffelSRLNMZLL20} is used to extract the text tokens. 
To calculate the combined embedding, for each pair of text token and image patch embeddings, it is determined whether the center of the text bounding box is inside of the image patch. If this is the case, the text features are added together with the image features of the patch. 
The resulting embeddings are concatenated with the image patches that do not contain any text. 
2D position embeddings are used as position information. 
While UDOP has an encoder-decoder architecture, we only use the encoder part and add a linear classifier head. 
During pretraining, UDOP incorporates curriculum learning to train from low-resolution to high-resolution images by scaling up the resolution in three steps: from 224 to 512, and then to 1024. 
Compared to LayoutLMv3, the base model of UDOP has more trainable parameters, approximately 794M. 
The encoder is a 24-layer transformer with 16 self-attention heads, hidden size of $D = 1,024$, and a feed-forward network with one hidden layer of size 4,096.

\subsection{Data Augmentation and Balancing}
\paragraph{Data Augmentation}

To increase the robustness of our models, data augmentation is applied to the two modalities of image and text. For the image modality, augmentations that adjust the noise (e.\,g., salt-and-pepper and Gaussian noise), brightness, color, and rotation of the image are randomly applied.
These augmentations are chosen based on the experience in improving the quality of Tesseract, a common open-source Optical Character Recognition (OCR) system.\footnote{\url{https://tesseract-ocr.github.io/tessdoc/ImproveQuality.html}} 
The degree to which the image is rotated is uniformly selected from within the range of -30 to 30 degrees. 
When rotation is applied to the image, the bounding box coordinates of the text are modified accordingly.

Three types of text data augmentation are implemented. 
The first is inserting single characters and the second is substituting single characters. 
\citet{HamdiPSCD23} studied how OCR errors affect NLP tasks of named entity recognition and named entity linking, and found that character insertion and substitution errors are hard to overcome. 
The third is deleting the first few characters of a text element. 
This is inspired by the fact that misidentifying the first character of a word along with incorrect word segmentation decreased the performance on NLP tasks~\cite{HamdiPSCD23}. 
This suggests that the first part of a word is more important than the latter part. 
The deletion length is uniformly selected from the range of 1 to 5 characters and applied to text elements with at least 10 characters. 

\paragraph{Data Balancing}

\begin{figure}[h]
  \centering
  \includegraphics[width=\linewidth]{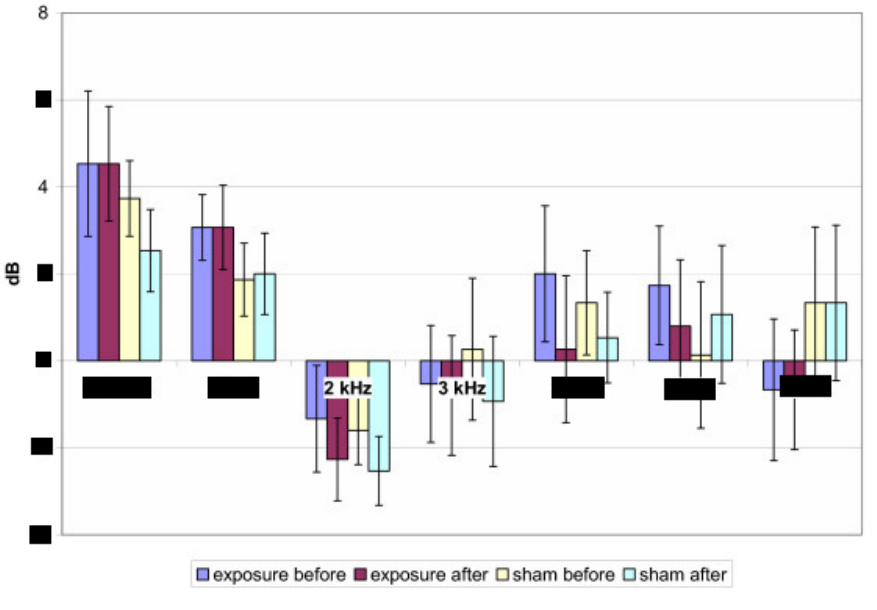}
  \caption{Demonstration of cutout augmentation applied to a bar chart from ICPR22. In this example, the chart is augmented with 10 masks for the tick label class.}
  \label{fig:figure-2}
\end{figure}

Due to the nature of charts, the text role class distribution of chart datasets is imbalanced. For example, most charts have axis titles and multiple tick labels as they provide basic information, making these text roles more common compared to the other text roles such as legend title and chart title. Table \ref{table:datasets-1} shows in detail how the chart datasets are imbalanced. Thus, to offset this data imbalance, two data balancing methods are implemented. The first method is a weighted cross-entropy loss function. 
The objective of this loss function is to assign more penalty to the minority class samples~\cite{RezaeiMF20}. 
The second data balancing method is modified cutout augmentation, which randomly masks regions of the image during training~\cite{DevriesT17}. 
The modified cutout augmentation randomly selects a text role class to be masked based on the class distribution.
For this class, our cutout augmentation chooses a random number of labels to mask. 
Using the bounding box coordinate information of the text, the area of the text in the chart image is masked. 
The resulting augmented chart has $n$ number of masks for the selected class. 
An example of modified cutout augmentation can be found in Figure \ref{fig:figure-2}.

\section{Experimental Apparatus}
\label{sec:experimentalapparatus}

We introduce the datasets used for finetuning and evaluating LayoutLMv3 and UDOP for text role classification. Next, we describe the procedure of our experiments and how we optimized the hyperparameters of the two models. Lastly, we explain the metric used to score the performance of the models.

\subsection{Datasets}
\label{sec:datasets}

A summary of all of the datasets can be found in Table \ref{table:datasets-2}. For all datasets, we assume that the text and bounding box coordinates are extracted accurately from each image.

\subsubsection{ICPR22}
\label{sec:ICPR22}

\begin{table*}[h!]
\centering
 \begin{tabular}{l|r|r|r|r|r}
 \hline
 \toprule
Text Role Class & ICPR22 (Train+Val.) & ICPR22 (Test) & CHIME-R  & DeGruyter & EconBiz \\ 
\midrule
Legend title & 190 & 31 & 1 & 2 & 7 \\
Chart title & 493 & 41 & 67 & 18 & 36 \\
Tick grouping & 792 & 57 & 5 & 79 & 36 \\
Mark label & 1,920 & 84 & 0 & 338 & 175 \\
Value label & 7,649 & 228 & 111 & 25 & 175 \\
Axis title & 10,721 & 613 & 140 & 289 & 125 \\
Legend label & 12,286 & 954 & 55 & 98 & 258 \\
Tick label & 95,430 & 5,605 & 1,183 & 1,583 & 1,805 \\
Other & 6,305 & 546 & 25 & 519 & 440 \\  \hline
Total & 135,786 & 8,159 & 1,587 & 2,951 & 3,057 \\ \hline
Avg per chart & 20.99 & 21.41 & 13.80 & 24.59 & 25.26 \\
 \bottomrule
 \end{tabular}
 \caption{Text role class distributions of the datasets. The numbers represent the total number of text roles for the text elements in the dataset. The distribution for the ICPR22 train dataset is the full train dataset before the train/validation split. The class distribution for the ICPR22 test dataset is the same for ICPR22-N, see Section \ref{sec:ICPR22-N}.
 For CHIME-R, DeGruyter, and EconBiz, we consider both the train and test data.
 }
 \label{table:datasets-1}
\end{table*}

\begin{figure*}
\centering
\begin{subfigure}{0.3\linewidth}
 \includegraphics[width=\textwidth]{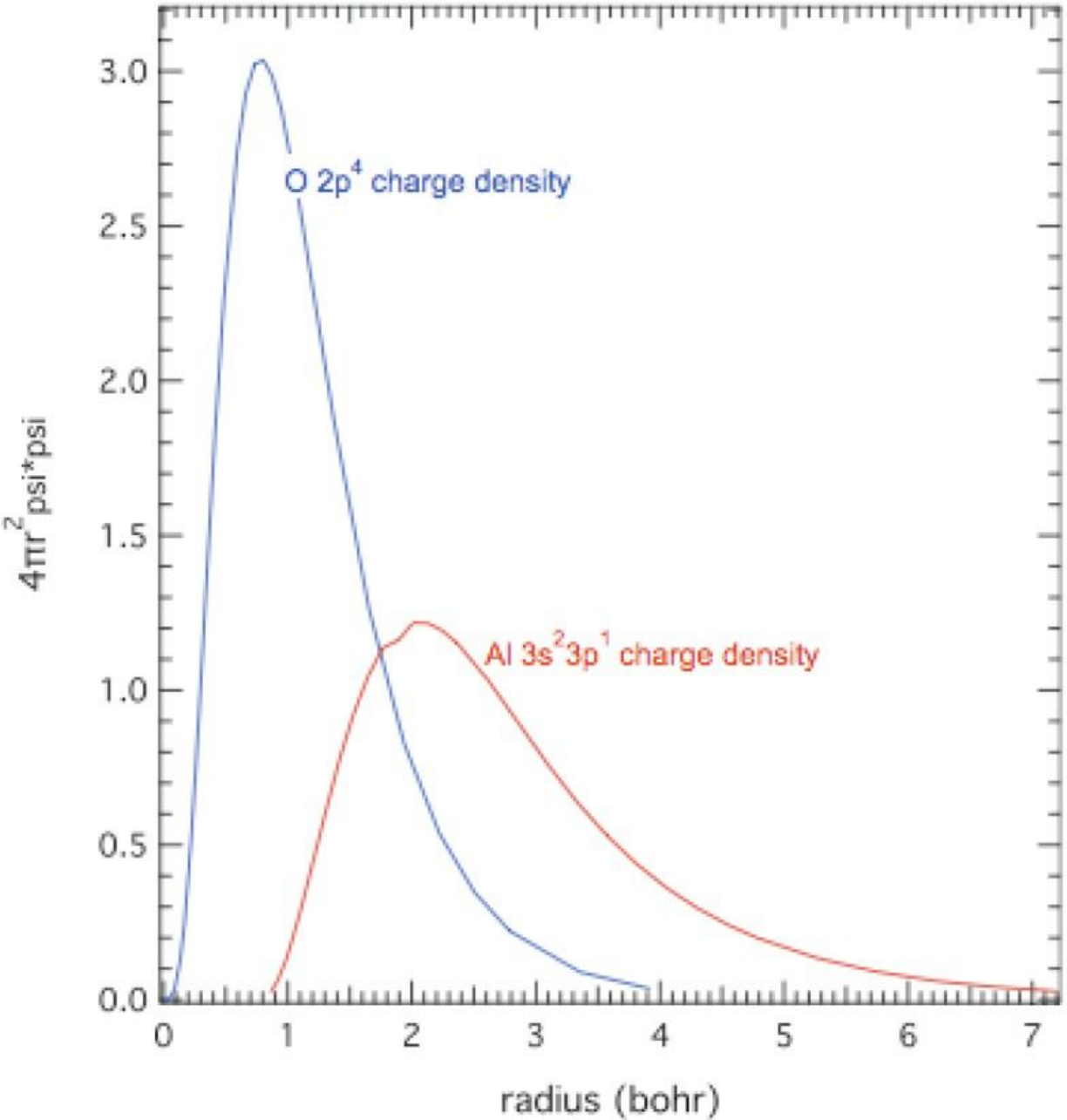}
 \caption{ICPR22}
\end{subfigure}
\hfill
\begin{subfigure}{0.3\linewidth}
 \includegraphics[width=\textwidth]{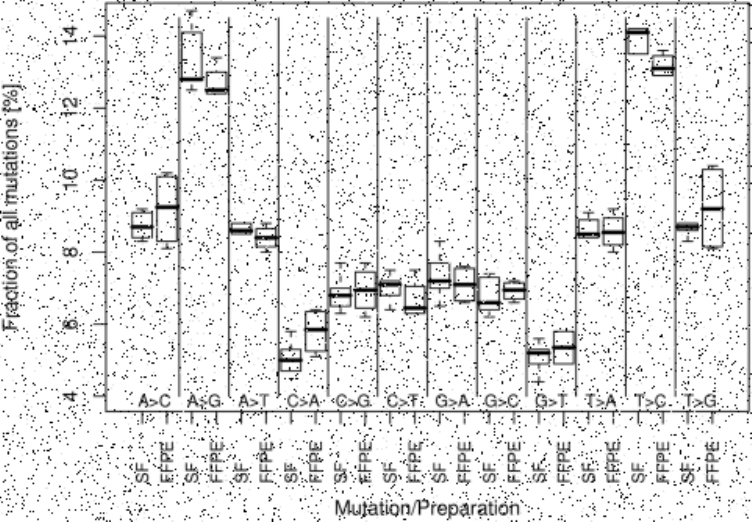}
 \caption{ICPR22-N}
\end{subfigure}
\hfill
\begin{subfigure}{0.3\linewidth}
 \includegraphics[width=\textwidth]{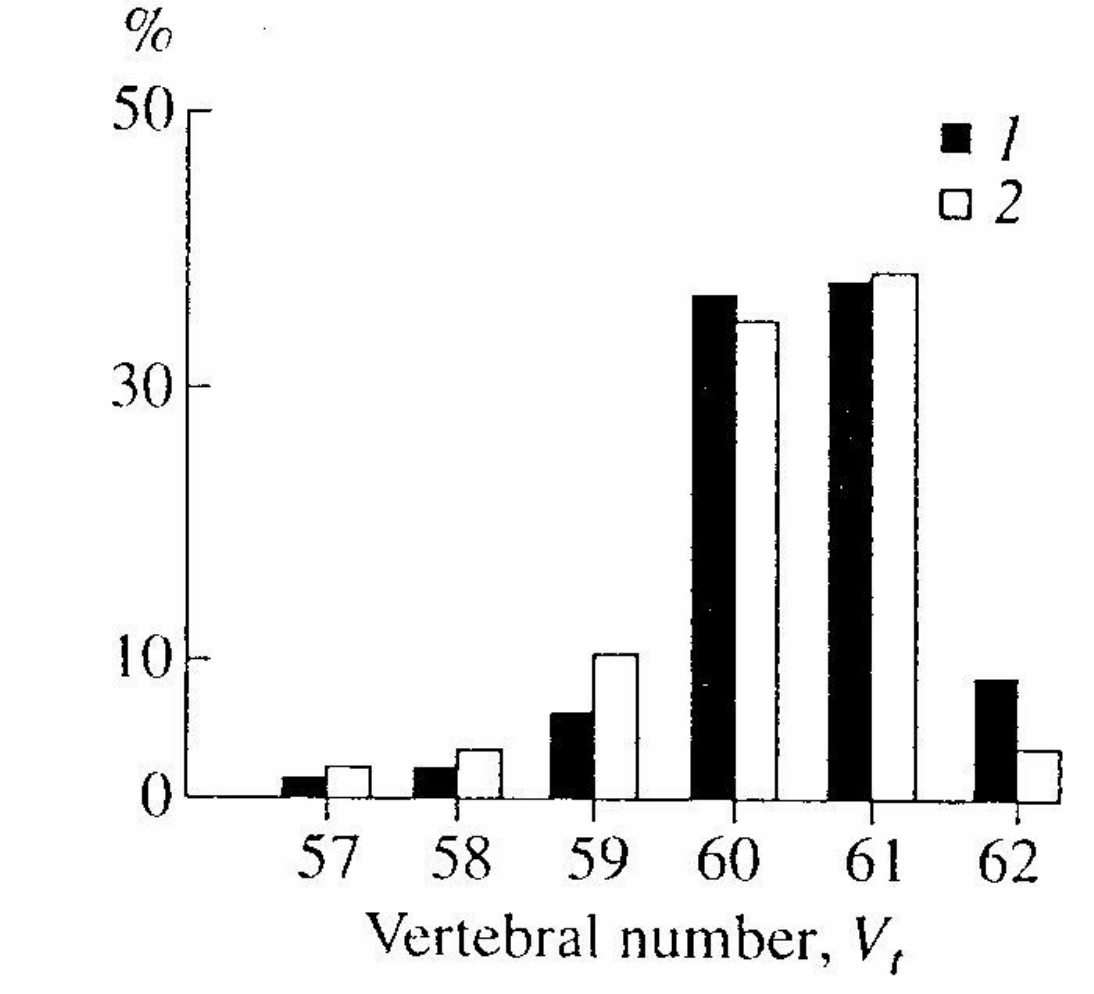}
 \caption{CHIME-R}
\end{subfigure}
\hfill
\begin{subfigure}{0.3\linewidth}
 \includegraphics[width=\textwidth]{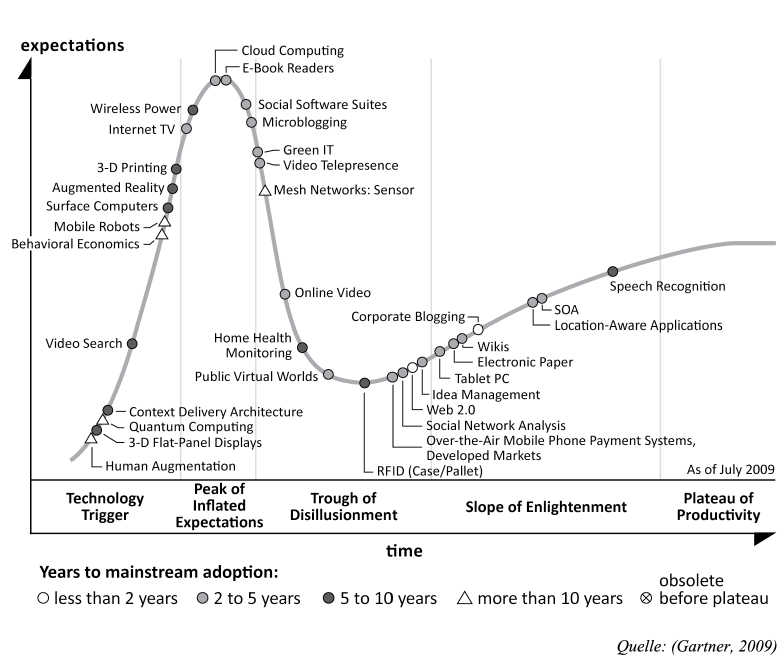}
 \caption{DeGruyter}
\end{subfigure}
\hspace{1.5cm}
\begin{subfigure}{0.3\linewidth}
 \includegraphics[width=\textwidth]{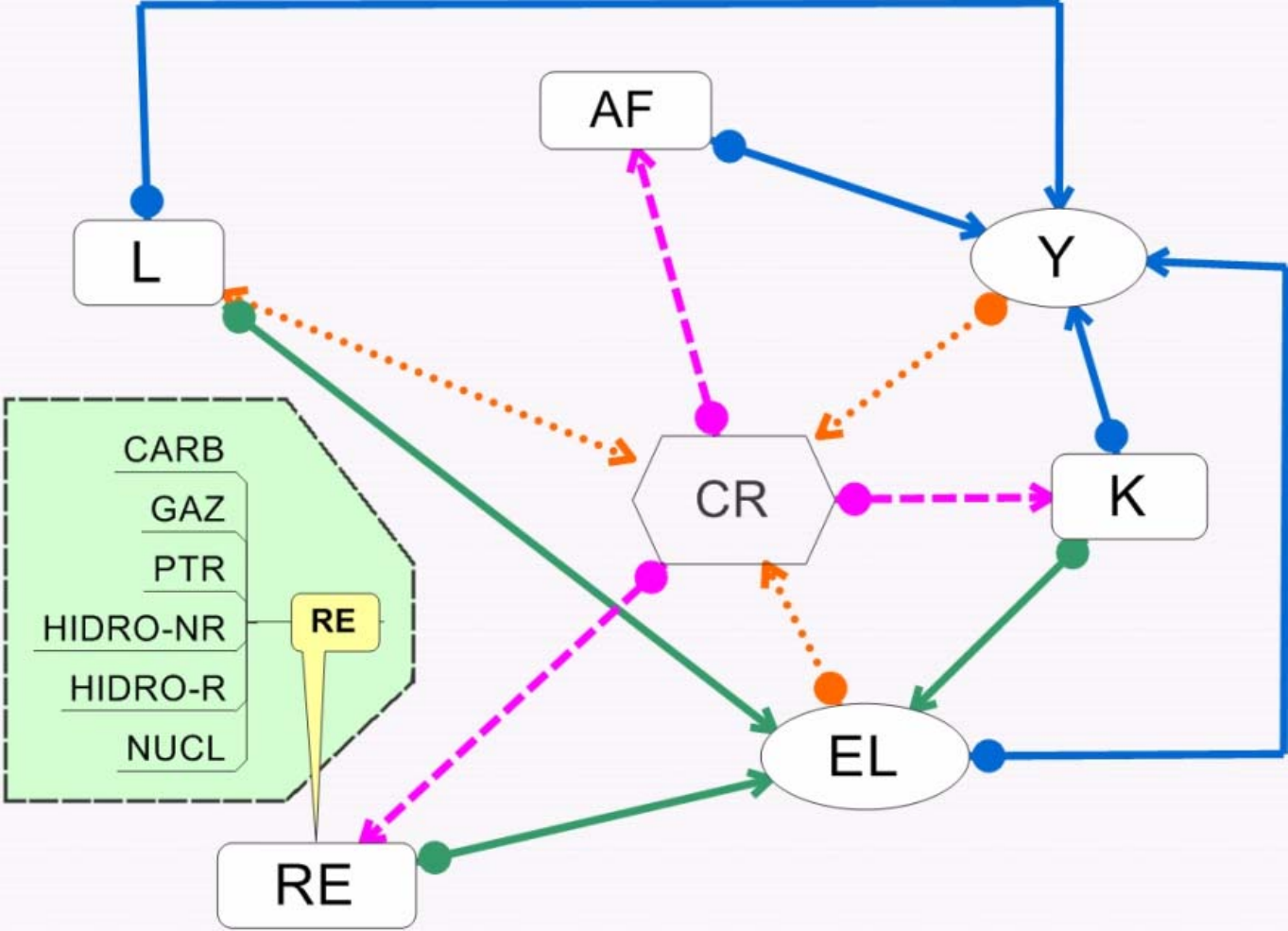}
 \caption{EconBiz}
\end{subfigure}
\caption{Example charts from each dataset}
\label{fig:examples}
\end{figure*}

This dataset is from the 2022 CHART-Infographics challenge~\cite{DavilaXAMSG22}. In previous years of the competition, two possible versions of the train and test datasets were provided: synthetic and real. The synthetic dataset consists of synthetic charts created from various online sources and the real dataset includes real charts extracted from the PubMedCentral (PMC) open-access repository. ICPR22 is the real PMC chart dataset from the ICPR 2022 challenge that includes a separate train dataset\footnote{\url{https://www.dropbox.com/s/85yfkigo5916xk1/ICPR2022_CHARTINFO_UB_PMC_TRAIN_v1.0.zip?dl=0}} and a test dataset\footnote{\url{https://www.dropbox.com/s/w0j0rxund06y04f/ICPR2022_CHARTINFO_UB_UNITEC_PMC_TEST_v2.1.zip?dl=0}}. While past synthetic chart data is available, only real charts are used because, in the past, training and testing on synthetic charts proved to be an easy task as most submission scores were perfect or near perfect~\cite{DavilaKSGTSC19, DavilaTSSSG20}. Furthermore, with the assumption that real charts are more complex than synthetic charts, training on synthetic charts does not provide additional benefits for minority class samples. For a given chart image in ICPR22, the chart type, text block bounding box coordinates, text transcription, text block ID, and text role label are provided. There are nine text role classes including chart title, legend title, legend label, axis title, tick label, tick grouping, mark label, value label, and other. Tick grouping is defined as the text that groups tick labels together. Value label is defined as text displaying quantitative data for a certain point or area on the chart. Mark label is defined as text that differentiates marks on charts (e.\,g., dotted line versus plain line). Figure \ref{fig:figure-1} shows an example of the classes tick grouping and value label, and the ICPR22 chart in Figure \ref{fig:examples} shows an example with the mark label class.

ICPR22 is preprocessed to only include samples that can be used for text role classification. 
This means that each sample must have an image and text, bounding box coordinates, and text role annotations. 
This limits the variety of chart types in the dataset to line, scatter, box, and bar plots. 
Additionally, preprocessing steps are also taken to handle negative bounding box coordinates. 
In some cases, the text was located on the edge of the image, resulting in negative coordinates. 
To make sure that all of the bounding box coordinates are within the boundary of the image, these negative coordinates are set to zero.

\subsubsection{ICPR22-N}
\label{sec:ICPR22-N}

After finetuning, the robustness of the models is tested on a synthetic but realistic noisy dataset ICPR22-N which we created from the ICPR22 test dataset by applying the data augmentation methods introduced in Section 3.2. 
Data augmentation is randomly selected and applied to each image from a pool of augmentations with equal probability. 
One type of augmentation is applied to each image.

\subsubsection{CHIME-R, DeGruyter, and EconBiz}
\label{sec:CHIME-R, DeGruyter, and EconBiz}

The generalizability of the models is also tested on three datasets from~\citet{BoschenBS18}.
CHIME-R~\cite{YangHL06} contains real figures originally collected by the Center for Information Mining and Extraction (CHIME) which includes bar, pie, and line charts. 
The DeGruyter dataset consists of figures from academic books by DeGruyter\footnote{\url{https://www.degruyter.com/}}. The figures in the EconBiz dataset~\cite{BoschenS15} were randomly extracted from a corpus of 288,000 open-access publications in the economics domain. The chart types included in the DeGruyter and EconBiz datasets are more complex than the CHIME-R dataset and include figures that are not charts (e.\,g., diagrams), charts without axes (e.\,g., flow charts), and figures with multiple subplots. 

Since the text roles are missing from CHIME-R, DeGruyter, and EconBiz in the gold standard created by~\citet{BoschenBS18}, the text roles are manually assigned to each text element using the same nine text role classes as ICPR22.

\subsection{Procedure}
\label{sec:procedure}

We finetune the pretrained models LayoutLMv3 and UDOP in their base model size. 
All of the layers are initialized with the pretrained weights except for the classification output layer. 
To compare the performance of LayoutLMv3 and UDOP on the ICPR22 test dataset, three baseline models are used: IIIT CVIT Chart Understanding, UB-ChartAnalysis, and CACHED. Details of the models are discussed in Section \ref{sec:textroleclassification}.

There are four stages of finetuning and evaluation. In the first stage, hyperparameter tuning is done only on the ICPR22 train dataset which is split into train and validation with a split of 0.9/0.1. After obtaining the best hyperparameters for each model, the models are evaluated on the full test datasets: ICPR22, ICPR22-N, CHIME-R, DeGruyter, and EconBiz. The ICPR22 validation dataset is only used in the first stage for hyperparameter tuning. 
The second stage is training on only the ICPR22 train dataset and additionally applying data augmentation and balancing during training. Afterwards, testing is done on each of the full test sets of the datasets. 
For the third and fourth stages, the CHIME-R, DeGruyter, and EconBiz datasets are split into train and test datasets with a split of 0.7/0.3. The resulting train splits are combined with the ICPR22 train dataset to create a combined train dataset. 

Furthermore, in the third stage, the models are trained on this combined train dataset without data augmentation and balancing, and evaluated on the test splits of all datasets. 
In the fourth stage, the models are trained on all datasets with data augmentation and balancing and then evaluated on the test splits of all datasets. 
When training on all datasets combined, ICPR22-N is excluded and only used for testing the robustness of the models as data augmentation is already applied to this noisy dataset.

\begin{table}[h!]
\centering
 \begin{tabular}{l|c|c|c}
 \hline
 \toprule
 Dataset & Train & Validation & Test \\ 
\midrule

 ICPR22 & 5,823 & 647 & 381 \\
 CHIME-R & 80 & - & 35 \\
 DeGruyter & 84 & - & 36 \\
 EconBiz & 84 & - & 37 \\
 \bottomrule
 \end{tabular}
 \caption{Number of images for all datasets in the train, validation, and test sets.
 When training using ICPR22, we test on all images from CHIME-R, DeGruyter, and EconBiz.}
 \label{table:datasets-2}
\end{table}

Before applying data augmentation and balancing, each method is tested individually on ICPR22 with LayoutLMv3, and only the best methods are used in the final setup. The results for selecting the data augmentation and balancing methods can be found in Appendix \ref{appendix:extendedresults}. For data augmentation, among the six proposed methods, three methods are used in the final setup: adjusting noise, deleting characters, and inserting characters. For data balancing, cutout augmentation is used. When data augmentation and balancing are applied, the augmented dataset is concatenated with the original dataset. 

\subsection{Hyperparameter Optimization}
\label{sec:hyperparameteroptimization}

For both models, the batch size and learning rate are optimized during finetuning. For UDOP, the further hyperparameters optimized include the warmup steps and weight decay. The hyperparameter settings reported in both LayoutLMv3 and UDOP to finetune each model on the FUNSD dataset are used as a starting point. The FUNSD dataset is a document dataset commonly used for named entity recognition which is similar to the task of text role classification. For LayoutLMv3, the batch size and learning rate are selected from the search space {16, 32, 64} and {1e-5, 2e-5, 3e-5, 4e-5, 5e-5}, respectively. Subsequently, for UDOP, warmup steps, batch size, learning rate, and weight decay are selected from the search space {1,000, 5,000}, {16, 32}, {1e-4, 2e-4, 3e-4, 4e-4, 5e-4}, and {1e-2, 1e-3, 1e-4}. Each model is trained for the same number of epochs, with a maximum of 20,000 steps for batch size 16, 10,000 steps for batch size 32, and 5,000 steps for batch size 64. All hyperparameter tuning results are reported in Appendix \ref{appendix:hyperparameteroptimization}.

The best hyperparameter combination for each model is used in the final setup. We finetune LayoutLMv3 for 10,000 steps with batch size 32 and learning rate 2e-5. For UDOP, we finetune for 20,000 steps with 1,000 warmup steps, batch size 16, learning rate 2e-4, weight decay 1e-2, and the Adam optimizer parameters $\beta_1 = 0.9$, and $\beta_2 = 0.98$.

\subsection{Metrics}
\label{sec:measures}

The main metric used to measure text role classification is the F1-macro score. This is the metric used in the ICPR 2022 competition. First, using precision and recall, the harmonic mean (F-measure) is calculated for each text role class. The final F1-macro score is the average of the per-class scores. For a benchmark comparison, the same evaluation method is used. Considering the data imbalance, the F1-micro score is reported as well. The F1-micro score is the average score over all samples, independent of the class.
We always report scores in percent.

\section{Results}
\label{sec:results}

For all stages of training and evaluation, LayoutLMv3 outperforms UDOP on all test datasets. 
The best-performing model on the ICPR22 test dataset is LayoutLMv3 trained on just ICPR22 with no data augmentation and balancing. 
The F1-macro score is 82.87, which is the highest score compared to the competition models. The F1-macro score for UDOP when only trained on ICPR22 with no data augmentation and balancing is 72.56, which is lower than the competition models. 
For both LayoutLMv3 and UDOP, the F1-micro scores surpass the F1-macro scores on all datasets in all stages of training and evaluation. 
Furthermore, both models show more robustness to the synthetic noisy dataset ICPR22-N compared to the real datasets CHIME-R, DeGruyter, and EconBiz. 
Among the real datasets, the F1 scores are highest for CHIME-R and lowest for DeGruyter.

\begin{table*}
\centering
\begin{tabular}{ll|ccccc} \toprule
    Model & \diagbox{Train data}{Test data} & ICPR22 & ICPR22-N & CHIME-R & DeGruyter & EconBiz  \\ \midrule
    \multirow{4}{*}{LayoutLMv3} & ICPR22 & \textbf{82.87} | 93.99 & \textbf{77.23} | \textbf{90.75} & 66.61 | 79.67 & 17.19 | 38.27 & 51.91 | 63.20 \\ 
    & ICPR22 with DAB & 81.55 | 93.66 & 77.02 | 90.31 & 58.87 | 76.11 & 19.58 | 39.52 & 51.19 | 65.86 \\
    & All datasets & 79.95 | 93.88 & 73.47 | 89.18 & 86.60 | 94.05 & 28.29 | 54.98 & 61.30 | 78.03 \\
    & All datasets with DAB & 82.48 | \textbf{94.31} & 75.75 | 90.72 & \textbf{95.60} | \textbf{97.62} & \textbf{44.44} | \textbf{58.37} & \textbf{66.40} | \textbf{78.60} \\ \hline
    \multirow{4}{*}{UDOP} & ICPR22 & 72.56 | 89.41 & 70.52 | 86.80 & 60.19 | 72.88 & 15.01 | 29.64 & 38.49 | 51.48 \\ 
    & ICPR22 with DAB & 71.24 | 88.85 & 68.92 | 87.04 & 58.84 | 71.96 & 15.32 | 30.49 & 36.78 | 51.67 \\
    & All datasets & 69.88 | 88.62 & 65.56 | 84.99 & 80.39 | 91.23 & 23.02 | 51.91 & \textbf{52.79} | \textbf{66.08} \\
    & All datasets with DAB & \textbf{76.22} | \textbf{91.00} & \textbf{71.97} | \textbf{88.38} & \textbf{89.44} | \textbf{95.76} & \textbf{33.79} | \textbf{58.67} & 48.11 | 61.41 \\
     \bottomrule
\end{tabular}
\caption{F1 scores (F1-macro | F1-micro) in $\%$ of LayoutLMv3 and UDOP are reported on various chart datasets. When a model is trained on only ICPR22, the model is evaluated on the full test dataset. When a model is trained on all datasets, the model is trained on the train split of all datasets, and evaluated on the test split of each dataset. DAB indicates that data balancing and augmentation methods are applied during training. The best F1-macro and F1-micro score of the models on each dataset is in bold.}
\label{table: results-1}
\end{table*}

For ICPR22-N, the best F1-macro score of 77.23 is from LayoutLMv3 trained on only ICPR22 with no data augmentation and balancing. 
LayoutLMv3 trained on all datasets with data augmentation and balancing produced the best F1-macro scores of 95.60, 44.44, and 66.40 for CHIME-R, DeGruyter, and EconBiz, respectively. 
The results of LayoutLMv3 and UDOP are summarized in Table \ref{table: results-1}, and the comparison to the challenge models on ICPR22 can be found in Table \ref{table: results-2}.

\begin{table}
    \centering
    \begin{tabular}{l|c} \toprule
        Model & F1-macro Score on ICPR22 \\ \midrule
        IIIT_CVIT_Chart_Understanding & 82.1 \\ 
        UB-ChartAnalysis & 73.6 \\ 
        CACHED & 78.7 \\ \hline
        LayoutLMv3 (ICPR22) & \textbf{82.87} \\ 
        UDOP (All with DAB) & 76.22 \\ 
         \bottomrule
    \end{tabular}
    \caption{F1-macro scores on ICPR22 between the best-performing models of LayoutLMv3 and UDOP and the baseline models.}
    \label{table: results-2}
\end{table}

In the second stage of training and evaluation, where the models are trained only on ICPR22 with data augmentation and balancing, both LayoutLMv3 and UDOP do not show any improvement on any dataset. 
However, in the third stage where the models are trained on all datasets, although there is no increase in F1 scores on ICPR22 and ICPR22-N, the F1 scores of both models improve for CHIME-R, DeGruyter, and EconBiz. 
When the models are trained on all datasets with data augmentation and balancing in the fourth stage, UDOP scores the highest F1-macro score on ICPR22 with a score of 76.22, which is 2.6 points higher than the baseline model UB-ChartAnalysis. 
On the other hand, for LayoutLMv3, the F1-macro score still remains the highest for ICPR22 when the model is trained on only ICPR22 with no data augmentation and balancing. 
However, LayoutLMv3 scores the highest F1-micro score of 94.31 on ICPR22 when it is trained on all datasets with data augmentation and balancing.

\begin{table}
    \centering
    \begin{tabular}{c|c} \toprule
        Training Steps & ICPR22 \\ \midrule 
        20,000 & 72.56 | 89.41 \\ 
        40,000 & 70.94 | 89.03 \\
        60,000 & 67.59 | 87.84 \\
        80,000 & 72.29 | 89.32 \\
        100,000 & \textbf{73.95} | \textbf{90.71} \\
         \bottomrule
    \end{tabular}
    \caption{The F1 scores (F1-macro | F1-micro) of UDOP on the ICPR22 test dataset with additional training steps.}
    \label{table: results-3}
\end{table}

\section{Discussion}
\label{sec:discussion}

First, we discuss the key results of our experiments. 
Next, the limitations are discussed with respect to the datasets and methods. 
Afterward, we explore possible avenues for future work.

\subsection{Key Results}
\label{sec:keyresults}

LayoutLMv3 performs better than UDOP on text role classification, and the best method on the ICPR22 test dataset is LayoutLMv3 trained on only ICPR22 with no data augmentation and balancing. 
Even when continuing training UDOP to 100,000 steps, the performance of the model only slightly increases and LayoutLMv3 still demonstrates superior performance. 
The results for training UDOP with additional steps can be seen in Table \ref{table: results-3}. 
LayoutLMv3 not only outperforms UDOP, but it is also computationally less expensive, requiring only a third of the training time of UDOP, which takes approximately 36 hours on one Tesla V100-SXM2-32GB to train for 20,000 steps. 
Although data augmentation and balancing methods improved the F1-micro scores on ICPR22 for both models, the F1-macro score only increased for UDOP.
Despite applying data balancing methods, the F1-micro score is always higher than the F1-macro score. 
This shows how much more difficult it is to get a high score for each class than a high score overall. 
Data imbalance with the smallest class occurring only $0.3\%$ as often as the largest class as can be seen in Table \ref{table:datasets-1}, remains a difficult problem to be solved.

For simple charts, i.\,e. charts with few textual elements, which are close to the training dataset distribution, pretrained document layout analysis models can be finetuned with high performance, even with a small dataset by applying data augmentation and balancing. 
Nevertheless, it is important to train on samples of the dataset as the performance of the models is low when this is not done.
However, for complex datasets that are further away from the train dataset distribution, data augmentation and balancing methods are enough to achieve good performance on text role classification. 

Furthermore, the robustness of the models to other datasets depends on the training dataset. 
When only trained on ICPR22, the models perform well on the synthetic noisy dataset ICPR22-N but struggle on the real chart datasets CHIME-R, DeGruyter, and EconBiz. 
Furthermore, the performance of the models decreases for more complex charts with more textual elements.
As shown in Table~\ref{table:datasets-1}, CHIME-R has the fewest elements and DeGruyter the highest.
However, when trained on all datasets with data augmentation and balancing, the performance of the models is better on CHIME-R, a real yet simple chart dataset, than ICPR22-N. 
This demonstrates that chart complexity is an important factor to consider when conducting a robustness test for text role classification. 
Apart from general chart complexity, it is possible that DeGruyter and EconBiz are farther away from the ICPR22 train dataset distribution compared to CHIME-R. With limited training data for DeGruyter and EconBiz, data augmentation methods may not have been enough to compensate for differences in their respective distributions.

\subsection{Limitations}
\label{sec:threattovalidity}

There are several limitations regarding the data augmentation methods and the datasets. First, the data augmentation methods applied are general augmentation methods that are not chart-specific. Furthermore, while we hoped that training on augmented text would make the models more robust to noise, this was not the case. Augmenting the text may have hindered learning the relationship between the text embedding and text role. 

\begin{figure}[h]
  \centering
  \includegraphics[width=\linewidth]{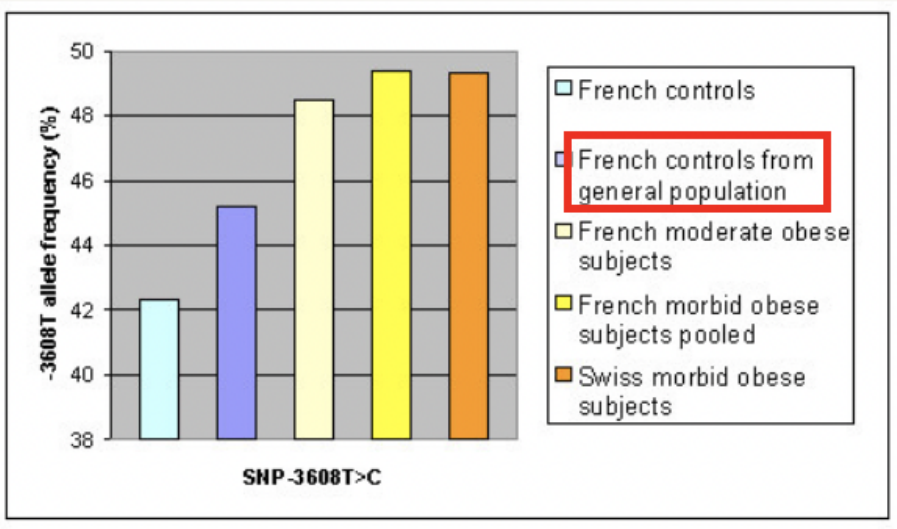}
  \caption{Example case where deleting characters from a text element resulted in character exclusions from the bounding box. Upon deleting "Fre" from the text element "French controls from general population" and adjusting the bounding box, "gen" in the following line is also excluded from the bounding box.}
  \label{fig:figure-4}
\end{figure}

Moreover, the bounding box coordinates are not adjusted when the data augmentation of deleting characters is applied. In some cases, text elements spanned more than one line in the image. Thus, deleting characters from the beginning of the text element and adjusting the bounding box coordinates afterward would result in the exclusion of characters from the following lines in the bounding box. An example of such a case can be found in Figure~\ref{fig:figure-4}. 
Adjusting the bounding boxes may be added in future work, but we do not expect that this will impact the results.

In addition, the text of the chart datasets is assumed to be in English. 
Therefore, it is unknown whether the models can be generalized to chart datasets in other languages. 
However, it is to be noted that the majority of scientific publications is in English. 
Finally, the text roles differ depending on the chart type. 
Thus, the generalizability of the models may have been influenced by the varying distribution of chart types within the datasets.

\subsection{Future Work}
\label{sec:futurework}

The results of our experiments can be generalized to other multimodal language tasks. 
As LayoutLMv3 outperformed UDOP in text role classification, the pretraining objective Word-Patch Alignment used in LayoutLMv3 may be a better method for learning the representations for other language tasks in which the position of the text is considered. 
This can also be seen in other tasks, where LayoutLMv3 outperforms UDOP such as named entity recognition on the text-centric document dataset FUNSD~\cite{TangYWFLZZZB}. 

When testing the models, it is important to account for the overall complexity of the chart dataset. As DeGruyter and EconBiz posed to be quite challenging datasets for text role classification, it may be the case that these datasets are also challenging for other chart analysis tasks such as chart summarization. Additionally, the data augmentation and balancing methods used in this paper are only tested on text role classification. They may be beneficial for other chart analysis tasks as well. This may be addressed developing chart-specific data augmentation methods that maintain the integrity of the text representation and performing text role classification based on the chart type. 

Other future work includes testing the larger model sizes of LayoutLMv3 and UDOP since we experiment with the base model sizes of LayoutLMv3 and UDOP in this paper. Additionally, one may use other document analysis models on chart datasets for text role classification. Lastly, the chart datasets CHIME-R, DeGruyter, and EconBiz annotated with text role information can be used for other chart analysis tasks. 

\section{Conclusion}
\label{sec:conclusion}

Without the cost of pretraining the models on a large-scale chart dataset, pretrained document layout analysis models can be finetuned on chart datasets for text role classification. 
In this paper, we evaluated the two pretrained document layout analysis models, LayoutLMv3 and UDOP, on five datasets ICPR22, ICPR22-N, CHIME-R, DeGruyter, and EconBiz. For the datasets CHIME-R, DeGruyter, and EconBiz, we introduced new text role class labels. 
Comparing the results of the models on ICPR22, LayoutLMv3 outperforms UDOP and all challenge entries. 
The results on ICPR22-N reveal that LayoutLMv3 is more robust to noise than UDOP, and the results on CHIME-R, DeGruyter, and EconBiz show that LayoutLMv3 also generalizes better. 
Although the performance of the models is relatively high on the simple dataset CHIME-R, the models struggle on the more challenging datasets DeGruyter and EconBiz. This confirms that due to the complexity of real charts, text role classification remains a difficult problem.

\begin{acks}

The authors acknowledge support by the state of Baden-Würt\-tem\-berg through bwHPC.
We thank Falk Böschen for providing the region annotations of the CHIME-R, DeGruyter, and EconBiz datasets~\cite{BoschenBS18,BoschenS15}. 
This work is co-funded under the 2LIKE project by the German Federal Ministry of Education and Research (BMBF) and the Ministry of Science, Research and the Arts Baden-Württemberg within the funding line Artificial Intelligence in Higher Education. 

\end{acks}

\bibliographystyle{ACM-Reference-Format}
\bibliography{dsbda-references}

\appendix

\FloatBarrier

\section{Supplementary Materials}
\label{appendix:supplementarymaterials}

We first report the results that led to the selection of the data augmentation and balancing methods. We also report all of the hyperparameter optimization results for LayoutLMv3 and UDOP.

\subsection{Data Augmentation and Balancing}
\label{appendix:extendedresults}

The results for selecting the data augmentation and balance methods are provided in Table \ref{table: appendix-1}. To select the methods, each method is applied individually to LayoutLMv3 trained on only the ICPR22 train dataset. The best methods are selected in the final setup of data augmentation and balancing: noise adjustment, character deletion, and character insertion for data augmentation, and cutout augmentation for data balancing.

\begin{table}
    \centering
    \begin{tabular}{l|rr} \toprule
        Data Augmentation and Balancing & F1-macro & F1-micro \\ \midrule
        Color adjustment & 79.95 & 93.17 \\ 
        Noise adjustment & \textbf{81.87} & \textbf{93.65} \\ 
        Rotation & 79.10 & 92.89 \\
        Character deletion & \textbf{82.29} & \textbf{93.58} \\
        Character insertion & \textbf{81.24} & \textbf{93.94} \\
        Character substitution & 80.85 & 93.77 \\ \hline
        Cutout augmentation & \textbf{81.84} & \textbf{94.18} \\
        Weighted cross-entropy loss & 78.68 & 93.49 \\
         \bottomrule
    \end{tabular}
    \caption{The F1 scores of LayoutLMv3 on the ICPR22 test dataset with each data augmentation and balancing method applied individually.}
    \label{table: appendix-1}
\end{table}

\subsection{Hyperparameter Optimization}
\label{appendix:hyperparameteroptimization}
Table \ref{table:appx} shows the results for hyperparameter tuning LayoutLMv3, and the results for UDOP can be found in Table~\ref{table: appendix-3}. The results reported are on the ICPR22 evaluation dataset. For UDOP, only the results for batch size 16 are reported because increasing the batch size generally decreased the performance of the model.

\begin{table*}
\centering 
 \begin{tabular}{ccc|rr}
 \toprule
Batch Size & Training Steps & LR & F1-macro & F1-micro \\ 
\midrule
16 & 20,000 & 1e-5 & 84.77 & 95.77 \\
16 & 20,000 & 2e-5 & 83.18 & 95.65 \\
16 & 20,000 & 3e-5 & 84.42 & 95.27 \\
16 & 20,000 & 4e-5 & 78.72 & 94.00 \\
16 & 20,000 & 5e-5 & 78.14 & 93.98 \\
32 & 10,000 & 1e-5 & 83.54 & 95.83 \\
\textbf{32} & \textbf{10,000} & \textbf{2e-5} & \textbf{87.24} & \textbf{96.64} \\
32 & 10,000 & 3e-5 & 85.24 & 96.49 \\
32 & 10,000 & 4e-5 & 81.18 & 94.63 \\
32 & 10,000 & 5e-5 & 82.22 & 94.37 \\
64 & 5,000 & 1e-5 & 83.41 & 94.37 \\
64 & 5,000 & 2e-5 & 80.08 & 94.04 \\
64 & 5,000 & 3e-5 & 81.41 & 94.62 \\
64 & 5,000 & 4e-5 & 84.30 & 94.98 \\
64 & 5,000 & 5e-5 & 82.61 & 94.62 \\ 
 \bottomrule
 \end{tabular}
 \caption{Hyperparameter optimization results for LayoutLMv3. The hyperparameters with the best results (in bold) on the ICPR22 evaluation dataset are selected for the final setup.}
\label{table:appx}
\end{table*}

\begin{table*}
\centering
 \begin{tabular}{ccccc|rr}
 \hline
 \toprule
Batch Size & Warmup Steps & Training Steps & Learning Rate & Weight Decay & F1-macro & F1-micro \\ 
\midrule
16 & 1,000 & 20,000 & 1e-4 & 1e-2 & 74.72 & 91.31 \\
\textbf{16} & \textbf{1,000} & \textbf{20,000} & \textbf{2e-4} & \textbf{1e-2} & \textbf{81.05} & \textbf{93.95} \\
16 & 1,000 & 20,000 & 3e-4 & 1e-2 & 75.86 & 91.65 \\
16 & 1,000 & 20,000 & 4e-4 & 1e-2 & 76.36 & 91.75 \\
16 & 1,000 & 20,000 & 5e-4 & 1e-2 & 75.93 & 92.98 \\
16 & 1,000 & 20,000 & 1e-4 & 1e-3 & 75.33 & 92.54 \\
16 & 1,000 & 20,000 & 2e-4 & 1e-3 & 77.30 & 91.76 \\
16 & 1,000 & 20,000 & 3e-4 & 1e-3 & 79.79 & 92.57 \\
16 & 1,000 & 20,000 & 4e-4 & 1e-3 & 77.30 & 91.76 \\
16 & 1,000 & 20,000 & 5e-4 & 1e-3 & 72.14 & 91.41 \\
16 & 1,000 & 20,000 & 1e-4 & 1e-4 & 74.44 & 91.91 \\
16 & 1,000 & 20,000 & 2e-4 & 1e-4 & 80.33 & 93.55 \\
16 & 1,000 & 20,000 & 3e-4 & 1e-4 & 77.42 & 92.15 \\
16 & 1,000 & 20,000 & 4e-4 & 1e-4 & 77.22 & 92.65 \\
16 & 1,000 & 20,000 & 5e-4 & 1e-4 & 77.18 & 93.00 \\
16 & 5,000 & 20,000 & 1e-4 & 1e-2 & 72.97 & 91.80 \\
16 & 5,000 & 20,000 & 2e-4 & 1e-2 & 80.49 & 93.49 \\
16 & 5,000 & 20,000 & 3e-4 & 1e-2 & 75.75 & 93.31 \\
16 & 5,000 & 20,000 & 4e-4 & 1e-2 & 73.87 & 91.93 \\
16 & 5,000 & 20,000 & 5e-4 & 1e-2 & 73.81 & 92.57 \\
16 & 5,000 & 20,000 & 1e-4 & 1e-3 & 75.08 & 92.17 \\
16 & 5,000 & 20,000 & 2e-4 & 1e-3 & 80.78 & 93.88 \\
16 & 5,000 & 20,000 & 3e-4 & 1e-3 & 71.15 & 91.81 \\
16 & 5,000 & 20,000 & 4e-4 & 1e-3 & 78.41 & 92.05 \\
16 & 5,000 & 20,000 & 5e-4 & 1e-3 & 76.18 & 92.50 \\
16 & 5,000 & 20,000 & 1e-4 & 1e-4 & 77.90 & 92.79 \\
16 & 5,000 & 20,000 & 2e-4 & 1e-4 & 78.88 & 92.62 \\
16 & 5,000 & 20,000 & 3e-4 & 1e-4 & 76.48 & 92.58 \\
16 & 5,000 & 20,000 & 4e-4 & 1e-4 & 74.11 & 91.91 \\
16 & 5,000 & 20,000 & 5e-4 & 1e-4 & 76.35 & 93.22 \\
 \bottomrule
 \end{tabular}
 \caption{Hyperparameter optimization results for UDOP. The hyperparameters with the best results (in bold) on the ICPR22 evaluation dataset are selected for the final setup.}
 \label{table: appendix-3}
\end{table*}

\end{document}